\documentclass[final,1p,times]{elsarticle}

\usepackage{amssymb}

\usepackage{lineno}

\journal{Pattern Recognition}

\usepackage[table]{xcolor}
\usepackage{graphicx}
\usepackage{amssymb}
\usepackage{amsmath}
\usepackage{multirow}
\usepackage{tabularx}
\usepackage{dsfont}
\usepackage{url}
\usepackage{color}
\usepackage{amsthm}
\usepackage[ruled,linesnumbered]{algorithm2e}
\usepackage[export]{adjustbox}
\usepackage{booktabs}
\usepackage{subcaption}
\usepackage{longtable}
\usepackage{tabu}
\usepackage{makecell} 
\usepackage{sidecap} 
\usepackage{bm}

\usepackage[section]{placeins}

\usepackage{xparse}
\usepackage{blindtext}
\usepackage{morewrites}
\usepackage{wrapfig}
\usepackage{xkeyval}
\usepackage{tcolorbox}
\newwrite\authorbibfile
\AtBeginDocument{%
  \immediate\openout\authorbibfile=\jobname.aub%
}%
\AtEndDocument{%
\immediate\closeout\authorbibfile
\InputIfFileExists{\jobname.aub}{}{}
}%

\makeatletter
\define@key{authorbib}{scale}[1]{%
\def\AuthorbibKVMacroScale{#1}%
}
\define@key{authorbib}{wraplines}[10]{%
\def\AuthorbibKVMacroWraplines{#1}%
}
\define@key{authorbib}{imagewidth}[4cm]{%
\def\AuthorbibKVMacroImagewidth{#1}%
}
\define@key{authorbib}{overhang}[10pt]{%
\def\AuthorbibKVMacroOverhang{#1}%
}
\define@key{authorbib}{imagepos}[l]{%
\def\AuthorbibKVMacroImagepos{#1}%
}
\makeatother

\presetkeys{authorbib}{imagepos=l, imagewidth=4cm, wraplines=8, overhang=20pt}{}
\newlength{\AuthorbibTopSkip}
\newlength{\AuthorbibBottomSkip}
\NewDocumentCommand{\authorbibliography}{+o+m+m+m}{%
  \IfNoValueTF{#1}{%
  }{%
    \setkeys{authorbib}{#1}%
    \immediate\write\authorbibfile{%
      \string\begin{wrapfigure}[\AuthorbibKVMacroWraplines]{\AuthorbibKVMacroImagepos}[\AuthorbibKVMacroOverhang]{\AuthorbibKVMacroImagewidth}^^J
        \string\includegraphics[scale=\AuthorbibKVMacroScale]{#2}^^J
        \string\end{wrapfigure}^^J
    }%
  }%
  \IfNoValueTF{#3}{%
    \typeout{Warning: No author name}%
  }{%
      \immediate\write\authorbibfile{%
      \unexpanded{\vspace{\AuthorbibTopSkip}}^^J
      \string\noindent\relax
      \unexpanded{\textbf{#3}\par}^^J
      \string\noindent\relax
      \unexpanded{#4}^^J%
      \unexpanded{\vspace{\AuthorbibBottomSkip}}^^J
      }%
  }%
}%

\definecolor{dg}{rgb}{0,0.694,0.298}
\definecolor{purple}{rgb}{0.4,0.176,0.569}
\definecolor{royalblue}{RGB}{65,105,225}
\definecolor{gray}{RGB}{100,100,100}

\usepackage{pifont}
\newcommand{\figref}[1]{Figure~\ref{#1}}
\newcommand{\reqref}[1]{Eq.~\eqref{#1}}
\newcommand{\secref}[1]{Section~\ref{#1}}
\newcommand{\tableref}[1]{Table~\ref{#1}}

\makeatletter
\DeclareRobustCommand\onedot{\futurelet\@let@token\@onedot}
\def\@onedot{\ifx\@let@token.\else.\null\fi\xspace}
\def\eg{\emph{e.g}\onedot} 
\def\ie{\emph{i.e}\onedot} 
 
\def\etc{\emph{etc}\onedot} 
 
\def\etal{\emph{et al}\onedot}
\makeatother

\definecolor{americanrose}{rgb}{1.0, 0.01, 0.24}

\newcommand{\stdc}[1]{\textbf{\textcolor{gray}{#1}}}

\newcommand{\revised}[1]{{#1}}

\usepackage{multirow}
\usepackage{colortbl}
\definecolor{tabgray}{rgb}{0.95,0.95,0.95}
\definecolor{tabhigh}{rgb}{0.90,0.90,0.90}
\definecolor{top1}{rgb}{1.0, 0.6, 0.6} 
\definecolor{top2}{rgb}{0.98, 0.91, 0.71}
\definecolor{top3}{rgb}{0.91, 1.0, 1.0}
\definecolor{top1-2}{rgb}{1.0, 0.66, 0.66} 
\definecolor{top1-3}{rgb}{1.0, 0.72, 0.72} 
\definecolor{top1-4}{rgb}{1.0, 0.78, 0.78} 
\definecolor{top1-5}{rgb}{1.0, 0.84, 0.84} 
\definecolor{top1-6}{rgb}{1.0, 0.90, 0.90} 
\definecolor{top1-7}{rgb}{1.0, 0.96, 0.96} 

\definecolor{citecolor}{RGB}{65,105,225}
\usepackage[breaklinks=true,colorlinks,citecolor=citecolor,bookmarks=false]{hyperref}

\begin{document}

\begin{frontmatter}

\title{\emph{DARTSRepair}:~{Core-Failure-Set Guided DARTS for Network Robustness to Common Corruptions}}

\author[inst1]{Xuhong Ren \corref{cor1}}
\author[inst2]{Jianlang Chen \corref{cor1}}
\author[inst6]{Felix Juefei-Xu}
\author[inst1]{Wanli Xue\corref{cor2}}
\author[inst3]{Qing Guo}
\author[inst4,inst2,inst5]{Lei Ma}
\author[inst2]{Jianjun Zhao}
\author[inst1]{Shengyong Chen}

\cortext[cor1]{Xuhong Ren and Jianlang Chen are co-first authors and contribute equally.}
\cortext[cor2]{Corresponding author (xuewanli@email.tjut.edu.cn).}

\affiliation[inst1]{organization={School of Computer Science and Engineering, Tianjin University of Technology},
city={Tianjin},
country={China}}
\affiliation[inst2]{organization={Kyushu University},
            country={Japan}}
\affiliation[inst6]{organization={Alibaba Group},
            country={USA}}
\affiliation[inst3]{organization={Nanyang Technological University},
country={Singapore}}
\affiliation[inst4]{organization={University of Alberta},
            country={Canada}}
\affiliation[inst5]{organization={Alberta Machine Intelligence Institute},
country={Canada}}
            

\begin{abstract}
%
Network architecture search (NAS), in particular the differentiable architecture search (DARTS) method, has shown a great power to learn excellent model architectures on the specific dataset of interest. 
In contrast to using a fixed dataset, in this work, we focus on a different but important scenario for NAS: how to refine a deployed network's model architecture to enhance its robustness with the guidance of a few collected and misclassified examples that are degraded by some real-world unknown corruptions having a specific pattern (\eg, noise, blur, \etc.).
To this end, we first conduct an empirical study to validate that the model architectures can be definitely related to the corruption patterns.
Surprisingly, by just adding a few corrupted and misclassified examples (\eg, $10^3$ examples) to the clean training dataset (\eg, $5.0\times10^4$ examples), we can refine the model architecture and enhance the robustness significantly.
To make it more practical, the key problem, \ie, how to select the proper failure examples for the effective NAS guidance, should be carefully investigated.
Then, we propose a novel \textit{core-failure-set guided DARTS} that embeds a $K$-center-greedy algorithm for DARTS to select suitable corrupted failure examples to refine the model architecture. 
We use our method for DARTS-refined DNNs on the clean as well as $15$ corruptions with the guidance of four specific real-world corruptions. 
Compared with the state-of-the-art NAS as well as data-augmentation-based enhancement methods, our final method can achieve higher accuracy on both corrupted datasets and the original clean dataset. On some of the corruption patterns, we can achieve as high as over $45\%$ absolute accuracy improvements.
\end{abstract}


\end{frontmatter}


\section{Introduction}
\label{sec:sample1}
\begin{figure}[t]
	\centering
	\includegraphics[width=1.0\columnwidth]{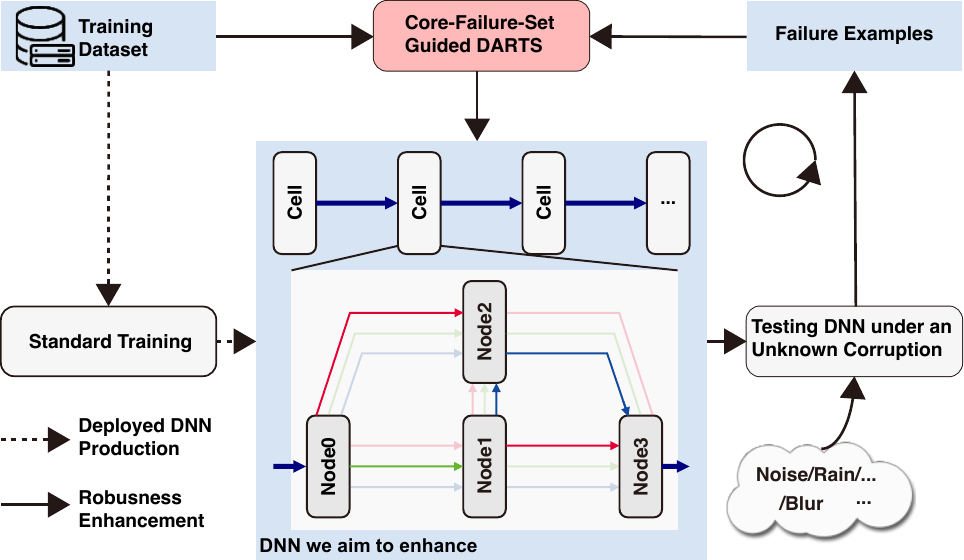}
	\caption{Pipeline of the proposed core-failure-set guided DARTS (\ie, CF-DARTS). The intuitive idea is to use a few collected failure examples to guide the network architecture search process to enhance the deployed DNN's robustness.
	}
	\label{fig:pipeline}
\end{figure}
%
In the process of neural network design, different tasks usually require the guidance of human experts' prior domain knowledge and repeated trial and error to obtain a high-performance model. This makes the design cost of a good neural network architecture rather expensive.
The emergence of neural architecture search (NAS) makes it possible to automate such a design process. As a classic NAS algorithm, differentiable architecture search (DARTS) \cite{liu2019darts}, while improving the performance of the model, reduces the architecture search time from thousands of GPU days \cite{NASNet,AmoebaNet} to 1.5 GPU days on the CIFAR-10 dataset. 
In many of the recent NAS-related studies, the focus has mostly been put on how to improve the model accuracy (most commonly for image classification tasks) \cite{tan2019efficientnet,pham2021meta,wongpanich202083} as well as model training and searching efficiency \cite{dong2020autohas,guo2020breaking,yan2019hm,stamoulis2019single}. However, one important aspect of an ideal DNN may be slightly overlooked, and that is the robustness improvement of the DNN models as a result of NAS. The DNN model robustness is a vitally important characteristic, especially for models that are deployed in the safety- and mission-critical environments. 

It is a known fact that (almost) all the high-performance DNN models that are naively trained with standard training datasets might suffer from robustness issues easily \cite{imagecorr}.
Specifically, they are prone to make erroneous decisions when presented with less ideal data (\eg, data that are perturbed by common real-world but unknown in advance corruption patterns), letting alone the examples that are adversarially crafted.

Therefore, in this paper, we will take a deeper dive into studying the robustness aspect of NAS. In particular, we take a special focus on studying how to refine a deployed DNN model's architecture for enhancing its robustness with the guidance of a few limited collected and misclassified examples that are degraded by some unknown but specific corruption patterns such as noise, blur, \etc, through the DARTS procedure. \revised{The intuitive idea and the whole pipeline are presented in \figref{fig:pipeline}.} 
As will be presented in a later section, our empirical study has validated that the model architectures are definitely related to the corruption patterns. We have made a surprising while interesting observation that by merely adding a few corrupted and misclassified examples (\eg, $10^3$ examples) into the clean training dataset (\eg, $5.0\times10^4$ examples), we can already refine the model architecture and significantly enhance the model robustness. 
For a more in-depth investigation of how to select the proper failure examples for the effective NAS guidance under a more practical setting, we propose a novel \textit{core-failure-set guided DARTS} that embeds a $K$-center-greedy algorithm for DARTS to select suitable corrupted failure examples to refine the model architecture. We have evaluated our method for DARTS-refined DNNs on the clean as well as $15$ corruptions with the guidance of four specific real-world corruptions, respectively. Compared with the state-of-the-art NAS methods as well as data-augmentation-based enhancement methods, our final method can achieve higher accuracy on all corrupted datasets and the original clean dataset. In particular, on some of the corruptions, we can achieve over $45\%$ absolute accuracy improvements. To the best of our knowledge, this work is among the very first attempts to investigate the novel aspect of DARTS for improving network robustness under the guidance of a few failure examples.

\section{Related Work}
\label{sec:relatedwork}

{\bf Network architecture search (NAS).}
Compared with manual architecture designing  \cite{AlexNet,ResNet,VGG}, NAS allows model designers to obtain a better model without professional domain knowledge and repeated trial and error. Early NAS algorithms are usually based on evolutionary algorithms \cite{AmoebaNet} and reinforcement learning \cite{NASNet}, which often require a lot of search time. DARTS \cite{liu2019darts} provides a new way of thinking, which relaxes the search space to make it continuous so that the searching process can be performed based on the gradient. This method can greatly reduce the time of architecture search so that the final architecture can be obtained within one GPU-day and has excellent performance.
Following DARTS, GDAS \cite{GDAS} proposes the differentiable architecture sampler, in which only one of the sub-graphs sampled need to be optimized at one training iteration, so as to reduce the search time efficiently. PC-DARTS \cite{PC-DARTS} proposes channel sampling and edge normalization technologies to solve the problem that DARTS requires large memory and computing when searching for models. P-DARTS \cite{P-DARTS} bridges the depth gap between the search net and the evaluation net in DARTS by gradually increasing the search depth. RobDARTS \cite{RobustDARTS} studies the failure mode of DARTS by looking at the largest eigenvalue Hessian matrix of the verification loss of the architecture and improves the robustness of DARTS based on the analysis. 
More recently, Tian~\etal \cite{tian2021108186} add a loss term for DARTS to alleviate the influence of the discretization of searching space. Hu~\etal \cite{HU2021108025} propose to reduce the degree of weight sharing of DARTS. As a result, the method benefits the more stable and accurate prediction. Guo~\etal \cite{GUO2022108448} accelerate the network architecture searching by avoiding the evaluation of candidate networks.
Xue~\etal \cite{xue2022108474} regard the automatic network architecture search as a combinatorial optimization problem on the search space and search strategies.
The above methods aim at improving the accuracy and search speed of DARTS, while our method intends to make the models searched from DARTS more robust.

{\bf Network robustness enhancement.}
The robustness of the deep neural network refers to the characteristic that the neural network can still maintain the normal input-output relationship when the input information or the neural network itself has limited disturbances. Through the adversarial attack algorithm, some adversarial samples with only a few differences can be automatically generated \cite{neurips20_abba,ijcai21_ava}, which makes the neural network make wrong judgments \cite{advatt}. Also, some common perturbations can also make the neural network make mistakes, such as pictures that are affected by overexposure, out of focus, bad weather, \etc \cite{hendrycks2019robustness}. 
To improve the robustness of the neural network, Hendrycks~\etal \cite{advatt2} propose parametric-noise-injection (PNI), which improves the model’s robustness against adversarial attacks and the accuracy of perturbed data by performing trainable Gaussian noise injection in activation and weights. Rusak~\etal \cite{imagecorr} use additive Gaussian and Speckle noise to adjust the training method and improve the performance of the model on common corruptions data. Schneider~\etal \cite{Batch} uses corrupted samples to correct the statistics of batch normalization and remove the covariate shift caused by common corruptions, thereby improving the DNN's accuracy for the corruption dataset. Compared with the above data-driven optimization methods, our method starts from a new angle and focuses on improving the model's robustness to corruption data by optimizing the model's architecture itself.

{\bf Subset selection methods.}
Our paper is inspired by active learning. For neural network training, due to the high cost of the manual labeling process, active learning is proposed, which can select a portion of the image from the collected data for annotation, thereby reducing the cost of labeling and still making the model with higher accuracy. Sener~\etal \cite{sener2018active} define active learning as a core-set selection problem. This is the first time that the core-set selection method has been applied to DNNs. Before this, the core-set selection method was applied to the core vector machine (CVM) \cite{coresetsvmcvm} to accelerate the training of SVM on large-scale data sets.  Har-Peled~\etal \cite{coresetkmeans} applied coreset to k-median and k-means clustering. Different from the above methods, it is an early exploration that we apply the core-set method to the refining of DARTS and improved the DNN's robustness to common corruptions.

\section{Methodology}

\begin{table}[t]
\centering
\tiny
\setlength{\tabcolsep}{1.0pt}
\caption{Preliminary experiment on CIFAR-10 dataset. A initial DNN's architecture is refined with the guidance of $\mathcal{C}_\text{fail}^{gn}$ and retrained on $\mathcal{D}_\text{train}$. $\mathcal{D}_\text{test}$ and $\mathcal{D}_\text{test}^{gn}$ are the original testing dataset of CIFAR-10 and its degraded counterpart via Gaussian noise.}
{
\begin{tabular}{c|c|c|c|c|c|c|c|c|c|c|c}
\toprule
\multicolumn{3}{c|}{Initial DNN}   
& \multicolumn{3}{c|}{DARTS-$10^{3}$GN}
& \multicolumn{3}{c|}{DARTS-$5\times10^{3}$GN}
& \multicolumn{3}{c}{DARTS-$10^{4}$GN} \\ \midrule
$\mathcal{D}_\text{test}$ & $\mathcal{D}_\text{test}^{\text{gn}}$ &  $\mathcal{D}_\text{fail}^\text{gn}/\mathcal{C}_\text{fail}^\text{gn}$
& $\mathcal{D}_\text{test}$ & $\mathcal{D}_\text{test}^{\text{gn}}$ &  $\mathcal{D}_\text{fail}^\text{gn}/\mathcal{C}_\text{fail}^\text{gn}$
& $\mathcal{D}_\text{test}$ & $\mathcal{D}_\text{test}^{\text{gn}}$ & 
$\mathcal{D}_\text{fail}^\text{gn}/\mathcal{C}_\text{fail}^\text{gn}$
& $\mathcal{D}_\text{test}$ & $\mathcal{D}_\text{test}^{\text{gn}}$  & 
$\mathcal{D}_\text{fail}^\text{gn}/\mathcal{C}_\text{fail}^\text{gn}$\\ \midrule
\multicolumn{1}{l|}{96.26} & \multicolumn{1}{l|}{32.58}  & 0.00 & \multicolumn{1}{l|}{96.73} & \multicolumn{1}{l|}{33.06}  & 14.06 & \multicolumn{1}{l|}{96.58} & \multicolumn{1}{l|}{35.57}  & 15.16 & \multicolumn{1}{l|}{96.77} & \multicolumn{1}{l|}{34.89}  & 11.82\\ \midrule
%

\end{tabular}
}
\label{tab:observation}
\end{table}

\subsection{Preliminary of DARTS}

Differentiable architecture search (DARTS) \cite{liu2019darts} is an impressive one-shot NAS method. It jointly optimizes the network's weights and model architectures represented by a supernet on the specific dataset of interest. 
As all model architectures are based on the basic building unit (\ie, cell) and share the network weights inherited from the supernet, they do not need to be retrained during the searching process, leading to efficient architecture search.
In contrast to the most of existing NAS-related works, our work explores how to use a few available failure examples to help robust network architecture search. 
For better understanding, we first review DARTS:

{\bf Architecture search space.}
DARTS is to search the architecture of the basic building unit, \ie, cell, and construct the whole network by stacking them.
In particular, each cell is a direct acyclic graph containing $N$ nodes and can be represented as $\mathcal{G}=(\mathcal{V},\mathcal{E})$. 
Each node $\text{V}_i\in \mathcal{V}$ corresponds to a feature map in the network while each edge $\text{E}_{i\to j}\in \mathcal{E}$ is an operation (\eg, $\text{O}_{i\to j}$) that maps the node $\mathrm{V}_i$ to another node $\mathrm{V}_j$.
The operation $\text{O}_{i\to j}$ is selected from an operation set $\mathcal{O}=\{\text{O}^{k}|k=1,\ldots,K\}$ including convolution, max pooling, average pooling,  identity, and zero representing for empty operation. 
Then, for each edge, we set a parameter $\alpha_k^{i\to j}$ to represent the contribution of the $k$th operation in $\mathcal{O}$ to a well-performed architecture.
The set $\bm{\alpha}=\{\alpha_k^{i\to j}|,i,j=1,\ldots,N,k=1,\ldots,K\}$ thus represents an architecture in the search space.
NAS is to optimize $\bm{\alpha}$ for higher performance.

{\bf Searching algorithm.} Given a training dataset $\mathcal{D}_\text{train}$ and a validation dataset $\mathcal{D}_\text{val}$, DARTS aims to solve two problems as follows,
\begin{align}
    \mathbf{W} &= \arg\min_{\mathbf{W}'} \mathcal{L}(\mathbf{W}',\bm{\alpha},\mathcal{D}_\text{train}), \text{~~and} \label{eq:darts_q1} \\
    \bm{\alpha} &= \arg\min_{\bm{\alpha}'} \mathcal{L}(\mathbf{W},\bm{\alpha}',\mathcal{D}_\text{val}), \label{eq:darts_q2}
\end{align}
where $\mathcal{L}(\cdot)$ is the cross-entropy loss function for classification. $\bm{\alpha}$ is a set of continuous variables representing the weights of all candidate operations in the supernet. DARTS uses the bilevel optimization to solve the two problems with approximated architecture gradients. The final architecture consists of the operations with the maximum weights in $\bm{\alpha}$ and the weights are retrained on the $\mathcal{D}_\text{train}$ instead of inheriting from the supernet. Please refer to \cite{liu2019darts} for more details. We finally represent a DNN by $\phi(\mathbf{W},\bm{\alpha})$ with $\mathbf{W}$ and $\bm{\alpha}$ for its weights and architecture, respectively.

\subsection{Problem Formulation and Empirical Study} \label{subsec:problem}
After training a deep model, it is highly desirable that it is robust to different corruptions that may happen in the real world.
Nevertheless, in practice, it is rather difficult to train such a perfect deep model due to the inevitable distribution shifting between the training dataset and real-world examples in the wild, as well as the less effective model architectures.
As a result, even state-of-the-art deep models do not always predict correctly when unknown or unseen specific corruptions appear \cite{hendrycks2019robustness}.
A practical and potentially feasible solution is to enhance the deep model's robustness with a few collected failure examples that may be degraded by specific corruptions, and enhance the model to be robust to similar corruption patterns while not harming its capability of handling the clean and other corrupted data.
Quite a few existing works address this problem via advanced data augmentation methods, \eg, AugMix \cite{hendrycks2020augmix} and CutMix \cite{Cutmix}, while ignoring the influence of model architectures.
In this work, we handle this problem from the angle by searching for better model architectures with the guidance of a few collected failure examples.
Note that, in contrast to the general robustness enhancement that can use pre-collected large-scale degraded examples to refine the model architecture, this task focuses on a more realistic scenario where the specific corruption pattern is unknown in advance and only a few collected failure examples are available. 
In the following, we give the definition of this task and provide some intuitive studies to reveal the feasibility and challenges of this idea.

{\bf Problem formulation.}
We first train a DNN $\phi(\mathbf{W},\alpha)$ on original datasets $\mathcal{D}_\text{train}$ and $\mathcal{D}_\text{val}$, and evaluate it on the original testing dataset $\mathcal{D}_\text{test}$. Then, we deploy it for real-world applications with the assumption the distribution of real-world examples is the same as the training dataset.
However, there can be numerous real-world corruptions \cite{hendrycks2019robustness} making the assumption invalid. 
When we fed $\phi$ with the degraded examples with a specific corruption, we inevitably get failure examples denoted as $\mathcal{D}_\text{fail}$.
Our goal is to refine the architecture of $\phi$ with the guidance of a small subset of $\mathcal{D}_\text{fail}$ (\ie, $\mathcal{C}_\text{fail}$), and enhance the model to generalize to similar corruption while not harming the accuracy on $\mathcal{D}_\text{test}$ and robustness to other corruptions.
We argue the reasons for using a small subset of $\mathcal{D}_\text{fail}$ for NAS as follows: 
\ding{182} A small subset of the failure examples would not result in extra high overhead for the process of the network architecture search, and thus it is more flexible for our method for real-world applications.
\ding{183} As analyzed in the following parts, with limited $\mathcal{D}_\text{fail}$, searching robust model architecture does not keep a positive relationship with the number of failure examples.

{\bf Observations.}
To understand the above problem clearly, we conduct a preliminary experiment on the CIFAR-10 dataset. Specifically, given a pre-trained deep neural network (DNN) optimized by DARTS on the CIFAR-10, we evaluate it on the testing examples corrupted by a specific corruption, \eg, Gaussian noise subset of CIFAR-10-C denoted as $\mathcal{D}_\text{test}^{\text{gn}}$  \cite{hendrycks2019robustness}.
As shown in the `initial DNN' column of \tableref{tab:observation}, we show that numerous testing examples are misclassified under this common corruption since the accuracy of original testing data reduces from $96.26\%$ to $32.58\%$.
We collect all failure examples as the set $\mathcal{D}_\text{fail}^\text{gn}$ having around $30,000$ examples.
Then, we randomly select $1,000$ failure examples from the $\mathcal{D}_\text{fail}^\text{gn}$ as the set $\mathcal{C}_\text{fail}^\text{gn}$. 
Moreover, we combine $\mathcal{C}_\text{fail}^\text{gn}$ and the original training dataset $\mathcal{D}_\text{train}$ as well as validation dataset $\mathcal{D}_\text{val}$ to refine the initial DNN's architecture via \reqref{eq:darts_q1} and \eqref{eq:darts_q2}.
After that, the architecture is fixed and retrained via $\mathcal{D}_\text{train}$.
Since the failure examples are only used in the architecture search stage, there is no overfitting risk on the model weights for those failure examples $\mathcal{C}_\text{fail}^\text{gn}$.
To avoid potential doubt, we compare the robustness of the original and refined DNNs on both $\mathcal{D}_\text{test}^\text{gn}$ and $\mathcal{D}_\text{fail}^\text{gn}$ without $\mathcal{C}_\text{fail}^\text{gn}$, \ie, $\mathcal{D}_\text{fail}^\text{gn}/\mathcal{C}_\text{fail}^\text{gn}$.
We also conduct this process with larger $\mathcal{C}_\text{fail}^\text{gn}$, \ie, $5\times 10^3$ and $10^4$ examples. All results are summarized in \tableref{tab:observation}.

Overall, we observe that: \ding{182} Even a few corrupted examples (\ie, $|\mathcal{C}_\text{fail}^\text{gn}|=1,000$ that is much smaller than $|\mathcal{D}_\text{train}|=50,000$) help search more robust model architecture to a specific corruption, that is, the accuracy on $\mathcal{D}_\text{fail}^\text{gn}$ increases from $0.0\%$ to $14.06\%$ while the accuracy on the original testing dataset (\ie, $\mathcal{D}_\text{test}$) becomes even higher. \ding{183} Different to the common understanding, the accuracy on $\mathcal{D}_\text{fail}^{\text{gn}}$ does not increase with the increasing of $|\mathcal{C}_\text{fail}^{\text{gn}}|$. 
We understand this can be due to the relationship between $\mathcal{C}_\text{fail}^\text{gn}$ and $\mathcal{D}_\text{train}$ as well $\mathcal{D}_\text{val}$ playing a key role for searching robust model architectures.
Hence, a more advanced core-failure-aware network architecture search method should be developed to select effective failure examples from $\mathcal{D}_\text{fail}^\text{gn}$.

\subsection{Core-Failure-Set-Guided DARTS}
\label{subsec:cfdarts}
According to the above analysis, we aim to identify a critical subset (\ie, $\mathcal{C}_\text{final}$) from the $\mathcal{D}_\text{fail}$ to make the searched model architecture be more robust to the specific corruption.
Specifically, given an initial DNN $\phi$ with its weights $\mathbf{W}$ and architecture $\bm{\alpha}$, we add an extra step before searching
\begin{align}\label{eq:darts_q0}
    \mathcal{C}_\text{fail} = \arg\min_{\mathcal{C}_\text{fail}'}~ & |\mathcal{L}(\phi,\mathcal{D}_\text{train}\cup\mathcal{D}_\text{fail})-\mathcal{L}(\phi,\mathcal{D}_\text{train}\cup\mathcal{C}_\text{fail}')|, \nonumber \\
    & \text{subject to}~~|\mathcal{C}_\text{fail}'|\leq B 
\end{align}
where $\mathcal{C}_\text{fail}$ is the selected failure examples for robust architecture searching and $B$ denotes the upper bound of the number of $\mathcal{C}_\text{fail}$.
Intuitively, \reqref{eq:darts_q0} is a core-set loss function \cite{sener2018active} that encourages the classification loss on $\mathcal{D}_\text{train}\cup\mathcal{C}_\text{fail}$ to be similar with that on $\mathcal{D}_\text{train}\cup\mathcal{D}_\text{fail}$, letting selected a few failure examples can cover situations of all failure examples when combining with $\mathcal{D}_\text{train}$.
Then, we split $\mathcal{C}_\text{fail}$ into two equal subsets denoted as $\mathcal{C}_\text{fail-t}$  and $\mathcal{C}_\text{fail-v}$, respectively. Moreover, the \reqref{eq:darts_q1} and \eqref{eq:darts_q2} for searching algorithm are modified as
\begin{align}
    \mathbf{W} &= \arg\min_{\mathbf{W}'} \mathcal{L}(\mathbf{W}',\bm{\alpha},\mathcal{D}_\text{train}\cup\mathcal{C}_\text{fail-t}), \text{~~and} \label{eq:darts_q1_new} \\
    \bm{\alpha} &= \arg\min_{\bm{\alpha}'} \mathcal{L}(\mathbf{W},\bm{\alpha}',\mathcal{D}_\text{val}\cup\mathcal{C}_\text{fail-v}). \label{eq:darts_q2_new}
\end{align}

Then, the key problem is how to solve \reqref{eq:darts_q0} effectively. Since not all labels of the examples in $\mathcal{D}_\text{fail}$ are given, we cannot naively solve \reqref{eq:darts_q0} via the gradient decent. Alternatively, as demonstrated in \cite{Wolf2011IJGIS} and \cite{sener2018active}, this problem has an upper bound and is equivalent to solving the K-center problem.

{\bf K-center greedy for $\mathcal{C}_\text{fail}$.} As demonstrated in \cite{Wolf2011IJGIS}, minimizing the core-set objective function is equivalent to the k-center problem, that is, choosing $B$ center examples to let the largest distance between an example in $\mathcal{D}_\text{train}$ and its nearest center be minimized. We can formulate it as
\begin{align}\label{eq:k-center}
    \mathcal{C}_\text{fail} = \arg\min_{\mathcal{C}_\text{fail}'}~ & \max_{\mathbf{X}_i\in\mathcal{C}_\text{fail}'}\min_{\mathbf{X}_j\in\mathcal{D}_\text{train}\cup\mathcal{C}_\text{fail}'} \text{dist}(\mathbf{X}_i,\mathbf{X}_j), \nonumber \\
    & \text{subject to}~~|\mathcal{C}_\text{fail}'|\leq B,
\end{align}
where $\text{dist}(\mathbf{X}_i,\mathbf{X}_j)$ denotes the distance between $\mathbf{X}_i$ and $\mathbf{X}_j$. Here, we adopt the feature of the layer before softmax of the initial DNN and their $L_2$-norm as the distance function. Although this problem is NP-hard, we can solve it efficiently via a greedy method \cite{sener2018active} and show the details in Algorithm~\ref{alg}.

\newcolumntype{h}{>{\columncolor{tabhigh}}c}
\begin{table*}[htbp]
\centering
\setlength{\tabcolsep}{1pt}
\tiny
\caption{Top-1 accuracy of the initial DNN and refined DNNs on the original CIFAR-10 testing dataset (\ie, `Clean' column), the specific corruption (\ie, Gauss$_\text{fail}$), and other 14 corruptions. DeepRepair, AugMix, and CutMix are data-augmentation-based robustness enhancements. AU-DARTS, RF-DARTS, and CF-DARTS are our methods based on the augmentation, random failure set selection, and core failure set selection strategies, respectively. AU/RF/CF-DARTSE retrains the refined architecture via the original training dataset and selected failure or augmented examples. All methods are evaluated five times, and the average accuracy, as well as the standard deviations, are reported in the first and second row of each method, respectively.}
{
\begin{tabular}{l|c|hcc|cccc|cccc|cccc}
\toprule 
\rowcolor{tabgray} &  & \multicolumn{3}{c|}{Noise} & \multicolumn{4}{c|}{Blur} & \multicolumn{4}{c|}{Weather} & \multicolumn{4}{c}{Digital}\tabularnewline
\midrule 
\rowcolor{tabgray} & Clean  & \cellcolor{tabhigh} Gauss$_\text{fail}$ & Shot  & Impulse  & Defocus  & Glass  & Motion  & Zoom  & Snow  & Frost  & Fog  & Bright  & Contrast  & Elastic  & Pixel  & Jpeg\tabularnewline
\midrule 
Org.~DNN  & 96.26  & 0.00  & 45.18  & 51.53  & 87.49  & 56.43  & 82.11  & 81.71  & 87.58  & 80.46  & 91.30  & 94.96  & 82.05  & 88.31  & 76.30  & 82.67\tabularnewline
\midrule 
DeepRepair & 95.25 & 29.26 & 46.09 & 40.83 & 80.66 & 50.56 & 78.08 & 78.07 & 79.09 & 81.51 & 90.31 & 88.63 & 87.35 & 76.12 & 66.11 & 63.69 \tabularnewline
& \stdc{${\pm0.86}$} & \stdc{${\pm7.06}$} & \stdc{${\pm4.99}$} & \stdc{${\pm3.93}$} & \stdc{${\pm0.85}$} & \stdc{${\pm3.48}$} & \stdc{${\pm1.36}$} & \stdc{${\pm1.09}$} & \stdc{${\pm0.90}$} & \stdc{${\pm0.94}$} & \stdc{${\pm0.41}$} & \stdc{${\pm0.68}$} & \stdc{${\pm0.58}$} & \stdc{${\pm0.92}$} & \stdc{${\pm0.82}$} & \stdc{${\pm1.37}$} \tabularnewline
%
AugMix & 89.75	& 50.35	& 67.58	& 71.6 & 87.62 & 61.06 & 83.98 & 85.98 & 82.81 & 81.28 & 89.88 & 90.18 & 89.61 & 82.73 & 69.22 & 73.90 \tabularnewline
& \stdc{${\pm1.02}$} & \stdc{${\pm3.11}$} & \stdc{${\pm1.74}$} & \stdc{${\pm2.54}$} & \stdc{${\pm1.48}$} & \stdc{${\pm2.00}$} & \stdc{${\pm1.81}$} & \stdc{${\pm1.61}$} & \stdc{${\pm2.37}$} & \stdc{${\pm3.51}$} & \stdc{${\pm1.38}$} & \stdc{${\pm1.65}$} & \stdc{${\pm1.39}$} & \stdc{${\pm1.17}$} & \stdc{${\pm2.56}$} & \stdc{${\pm1.79}$} \tabularnewline
CutMix & 95.70 & 21.99 & 53.82 & 50.61 & 84.36 & 62.58 & 80.60 & 78.22 & 85.52 & 81.80 & 89.35 & 94.06 & 77.63 & 85.54 & 75.49 & 79.04\tabularnewline
& \stdc{${\pm0.16}$} & \stdc{${\pm0.90}$} & \stdc{${\pm0.98}$} & \stdc{${\pm3.77}$} & \stdc{${\pm0.80}$} & \stdc{${\pm3.13}$} & \stdc{${\pm1.12}$} & \stdc{${\pm1.32}$} & \stdc{${\pm0.60}$} & \stdc{${\pm0.89}$} & \stdc{${\pm0.22}$} & \stdc{${\pm0.06}$} & \stdc{${\pm0.47}$} & \stdc{${\pm0.90}$} & \stdc{${\pm1.75}$} & \stdc{${\pm1.47}$} \tabularnewline
\midrule 
AU-DARTS & 97.15 & 17.19 & 50.99 & 50.87 & 88.09 & 67.90 & 84.86 & 81.31 & 88.58	& 84.07 & 91.76 & 94.04 & 83.06 & 88.63 & 78.62& 82.65 \tabularnewline
& \stdc{${\pm0.09}$} & \stdc{${\pm4.82}$} & \stdc{${\pm2.74}$} & \stdc{${\pm3.14}$} & \stdc{${\pm0.96}$} & \stdc{${\pm0.84}$} & \stdc{${\pm0.60}$} & \stdc{${\pm1.49}$} & \stdc{${\pm0.22}$} & \stdc{${\pm0.49}$} & \stdc{${\pm0.28}$} & \stdc{${\pm0.08}$} & \stdc{${\pm0.90}$} & \stdc{${\pm0.35}$} & \stdc{${\pm1.25}$} & \stdc{${\pm0.74}$} \tabularnewline
AU-DARTSE & 97.04 & 58.37 & 71.74 & 78.00 & 92.79 & 72.15 & 90.76 & 89.47 & 91.14	& 87.15 & 93.80 & 95.97 & 87.34 & 91.22 & 81.59 & 84.97 \tabularnewline
& \stdc{${\pm0.17}$} & \stdc{${\pm3.43}$} & \stdc{${\pm1.62}$} & \stdc{${\pm0.70}$} & \stdc{${\pm0.26}$} & \stdc{${\pm1.70}$} & \stdc{${\pm0.39}$} & \stdc{${\pm0.17}$} & \stdc{${\pm0.18}$} & \stdc{${\pm0.13}$} & \stdc{${\pm0.05}$} & \stdc{${\pm0.10}$} & \stdc{${\pm0.34}$} & \stdc{${\pm0.24}$} & \stdc{${\pm0.55}$} & \stdc{${\pm0.12}$} \tabularnewline
RF-DARTS  & 96.80  & 17.07  & 49.66  & 52.24   & 87.94   & 66.58  & 85.00   & 82.27  & 88.82  & 84.05  & 91.90  & 95.58  & 83.97  & 89.74  & 79.43  & 84.02 \tabularnewline
& \stdc{${\pm0.27}$} & \stdc{${\pm2.24}$} & \stdc{${\pm1.53}$} & \stdc{${\pm4.37}$} & \stdc{${\pm0.81}$} & \stdc{${\pm3.17}$} & \stdc{${\pm0.51}$} & \stdc{${\pm1.09}$} & \stdc{${\pm0.96}$} & \stdc{${\pm1.42}$} & \stdc{${\pm0.69}$} & \stdc{${\pm0.27}$} & \stdc{${\pm1.41}$} & \stdc{${\pm0.26}$} & \stdc{${\pm1.99}$} & \stdc{${\pm0.27}$} \tabularnewline
RF-DARTSE & 97.48 & 82.87 & 87.42 & 66.50 & 88.84 & 77.15 & 85.85 & 83.51 & 90.98 & 90.25 & 92.63 &	96.42 & 84.87 & 90.74 & 81.15 & 85.82 \tabularnewline
& \stdc{${\pm0.10}$} & \stdc{${\pm1.43}$} & \stdc{${\pm1.02}$} & \stdc{${\pm5.23}$} & \stdc{${\pm0.59}$} & \stdc{${\pm2.49}$} & \stdc{${\pm0.72}$} & \stdc{${\pm1.27}$} & \stdc{${\pm0.36}$} & \stdc{${\pm0.85}$} & \stdc{${\pm0.20}$} & \stdc{${\pm0.15}$} & \stdc{${\pm1.60}$} & \stdc{${\pm0.21}$} & \stdc{${\pm1.03}$} & \stdc{${\pm0.48}$} \tabularnewline
CF-DARTS  & 96.83   & 18.27  & 50.21  & 49.25   & 87.74  & 68.03   & 84.96  & 82.67  & 89.35   & 84.67   & 92.08   & 95.66   & 83.69  & 89.92   & 78.96  & 84.04\tabularnewline
& \stdc{${\pm0.20}$} & \stdc{${\pm2.86}$} & \stdc{${\pm1.62}$} & \stdc{${\pm4.03}$} & \stdc{${\pm0.32}$} & \stdc{${\pm2.76}$} & \stdc{${\pm0.37}$} & \stdc{${\pm0.83}$} & \stdc{${\pm0.39}$} & \stdc{${\pm1.13}$} & \stdc{${\pm0.36}$} & \stdc{${\pm0.21}$} & \stdc{${\pm0.76}$} & \stdc{${\pm0.20}$} & \stdc{${\pm1.86}$} & \stdc{${\pm0.94}$} \tabularnewline
CF-DARTSE  & 97.35 & 87.77 & 90.40 & 75.93 & 89.08 & 76.26 & 86.88 &84.21 & 91.01 & 89.66 & 92.98 & 96.39 & 84.71 & 91.00 & 80.62 & 85.28 \tabularnewline
& \stdc{${\pm0.10}$} & \stdc{${\pm2.08}$} & \stdc{${\pm0.83}$} & \stdc{${\pm5.93}$} & \stdc{${\pm0.74}$} & \stdc{${\pm2.51}$} & \stdc{${\pm0.51}$} & \stdc{${\pm1.45}$} & \stdc{${\pm0.50}$} & \stdc{${\pm1.21}$} & \stdc{${\pm0.39}$} & \stdc{${\pm0.09}$} & \stdc{${\pm0.91}$} & \stdc{${\pm0.22}$} & \stdc{${\pm0.59}$} & \stdc{${\pm0.62}$} \tabularnewline
\bottomrule
\end{tabular}}
\label{tab:baseline}
\end{table*}

\begin{algorithm}[tb]
\small
	{
		\caption{{$\mathcal{C}_\text{fail}$ selection via K-center greedy}}\label{alg}
		\KwIn{$\mathcal{D}_\text{train}$, the upper bound $B$, an initial DNN $\phi$ with its $\mathbf{W}_0$ and architecture $\bm{\alpha}_0$, and $\mathcal{D}_\text{fail}$.}
		\KwOut{$\mathcal{C}_\text{fail}$.}
		Initialize $\mathcal{C}_\text{fail}$ as empty \;
 		\For{$1\ \mathrm{to}\ B$}{
 		    $\mathbf{X}=\arg\max_{\mathbf{X}_i\in\mathcal{D}_\text{fail}}\min_{\mathbf{X}_j\in\mathcal{D}_\text{train}}~\text{dist}(\mathbf{X}_i,\mathbf{X}_j)$\;
 		    $\mathcal{C}_\text{fail}=\mathcal{C}_\text{fail}\cup\mathbf{X}$\;
    	}
	}

\end{algorithm}

\subsection{Implementation Details} \label{subsec:impl}

{\bf Dataset configuration.} Given a dataset including $\mathcal{D}_\text{train}$, $\mathcal{D}_\text{val}$ and $\mathcal{D}_\text{test}$, \eg, CIFAR-10 dataset \cite{cifar10}, we first train a model on $\mathcal{D}_\text{train}$ and $\mathcal{D}_\text{val}$ via DARTS as the initial DNN denoted as the $\phi_0$ with weights and architecture as $\mathbf{W}_0$ and $\bm{\alpha}_0$. 
Then, we build a degraded testing dataset by adding a specific corruption to $\mathcal{D}_\text{test}$, which is denoted as $\mathcal{D}_\text{test}^\text{xx}$ where $\text{xx
}$ is the name short for the corruption.
We evaluate the $\phi_0$ on $\mathcal{D}_\text{test}^\text{xx}$ and collect all failure examples as $\mathcal{D}_\text{fail}^\text{xx}$.
Our main goal is to enhance the robustness of $\phi_0$ for the corruption `$\text{xx}$' while not harming the effectiveness of handling other corruptions.

{\bf Evaluation configuration.} We use Algorithm~\ref{alg} to obtain the core failure set $\mathcal{C}_\text{fail}^\text{xx}$ from $\mathcal{D}_\text{fail}^\text{xx}$ with $B=1000$ examples included in $\mathcal{C}_\text{fail}^\text{xx}$. Then, we refine $\phi_0$ based on \reqref{eq:darts_q1_new} and \reqref{eq:darts_q2_new} and the datasets $\mathcal{D}_\text{train}\cup\mathcal{C}_\text{fail-t}$ and $\mathcal{D}_\text{val}\cup\mathcal{C}_\text{fail-v}$, and we can get a refined network $\phi_1$ with updated weights $\mathbf{W}_1$ and $\alpha_1$.
We denote the above process as an iteration to refine the targeted DNN $\phi_0$. Actually, we can further refine $\mathbf{W}_1$ and $\alpha_1$ for a second iteration refinement and get $\mathbf{W}_2$ and $\alpha_2$.
In practice, we only perform the refinement once for the weights and architecture optimization due to the computational expensive methods. 
In each iteration, the optimization of the architecture and weights requires several epochs (\ie, we set 20 epochs in our work) as done in the standard mini-batch-based training process. 
Specifically, in each epoch, we first randomly sample a batch from the training dataset (\ie, $\mathcal{D}_\text{train}\cup\mathcal{C}_\text{fail-t}$) and use the \reqref{eq:darts_q1_new} to update the network weights while fixing the architecture parameters $\alpha$. Then, we randomly sample a batch from the validation dataset (\ie, $\mathcal{D}_\text{val}\cup\mathcal{C}_\text{fail-v}$) and employ \reqref{eq:darts_q2_new} to optimize the architecture parameters while the weights stay the same. We conduct the above min-batch-based optimization for around 520 times for each epoch, and perform 20 epoches totally for $\phi_0$'s refinement.
We denote our method, \ie, \textit{core-failure-set guided DARTS}, as CF-DARTS.
Moreover, we implement a more advanced version of CF-DARTS by combining the core failure set $\mathcal{C}_\text{fail}^\text{xx}$ with $\mathcal{D}_\text{train}$ to retrain the DNN, and we denote this version as CF-DARTSE.
Finally, we evaluate and validate our method to observe the accuracy of refined model (\ie, $\bm{\alpha}_1$) on $\mathcal{D}_\text{fail}^\text{xx}/\mathcal{C}_\text{fail}^\text{xx}$ meaning the dataset $\mathcal{D}_\text{fail}^\text{xx}$ excluding $\mathcal{C}_\text{fail}^\text{xx}$, which is also represented as `xx$_\text{fail}$' in \tableref{tab:baseline}, \eg, `Gauss$_\text{fail}$' for $\mathcal{D}_\text{fail}^\text{gn}/\mathcal{C}_\text{fail}^\text{gn}$.

\section{Experimental Results}
\label{sec:exp}

\begin{figure}[t]
	\centering
	\includegraphics[width=1.0\columnwidth]{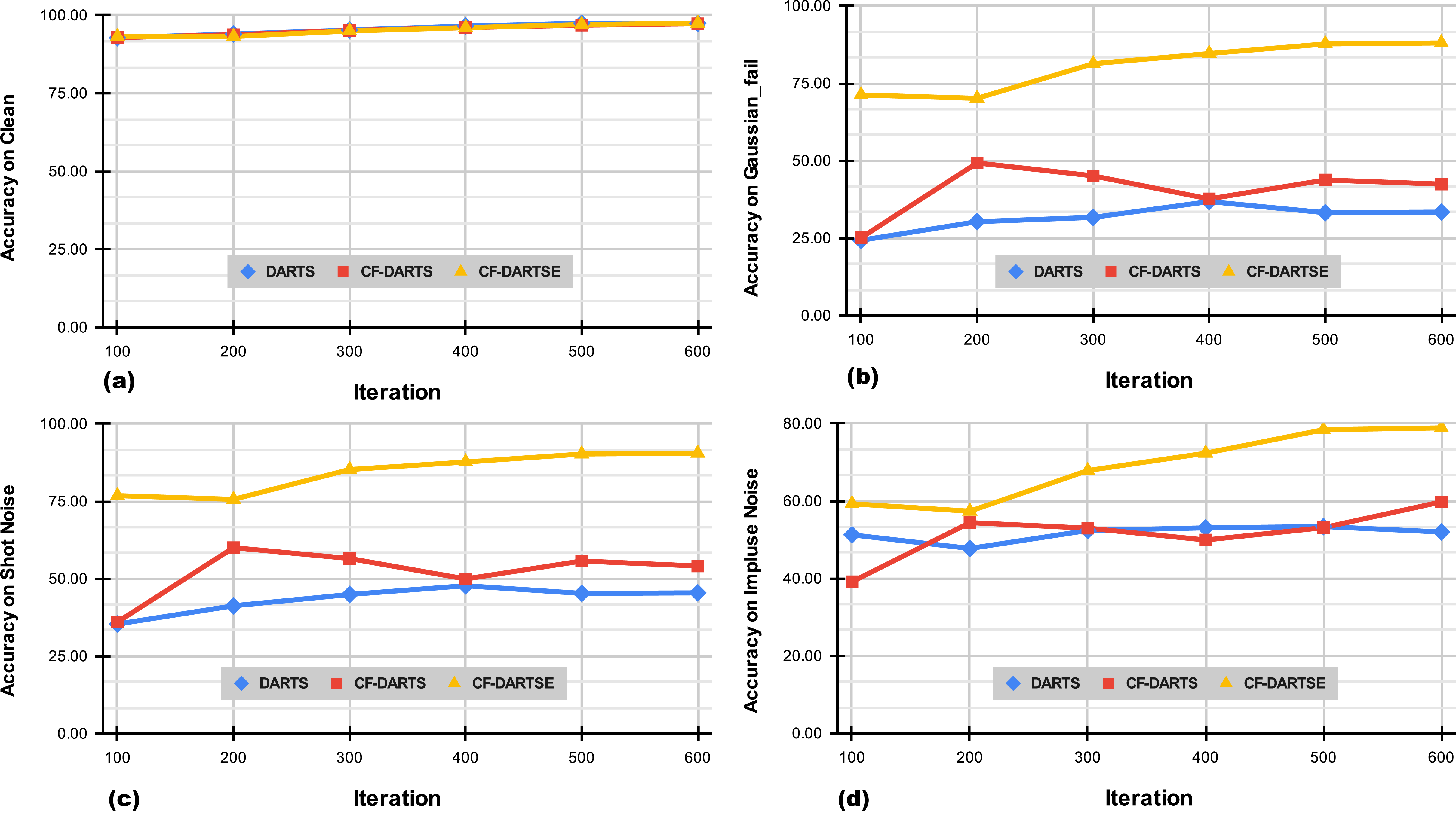}
	\caption{Accuracy of DARTS, CF-DARTS, and CF-DARTSE on the datasets of clean (\ie, (a)), Gaussian$_\text{fail}$ (\ie, (b)), shot noise (\ie, (c)), and impulse noise (\ie, (d)) along the training iterations.}
	\label{fig:budgecmp}
\end{figure}

\subsection{Setups}
\label{subsec:setups}
{\bf Datasets.}
Our experiments are performed on CIFAR-10~\cite{cifar10}, CIFAR-10-C~\cite{hendrycks2019robustness}, Tiny-ImageNet \cite{le2015tiny}, and Tiny-ImageNet-C \cite{hendrycks2019robustness}. CIFAR-10 consisting of $60,000$ 32x32 color images with 10 classes and each class has $6,000$ images, including $50,000$ training and $10,000$ testing images. CIFAR-10-C is an extension of CIFAR-10, adding $15$ corruptions to the original testing dataset of CIFAR-10. The corruption names are shown in \tableref{tab:baseline} and \ref{tab:different_number}. 
Similarly, we can also set up the Tiny-ImageNet \cite{le2015tiny} and Tiny-ImageNet-C datasets \cite{hendrycks2019robustness}. Then, we retrain the network architectures searched on CIFAR-10 and test them on Tiny-ImageNet and Tiny-ImageNet-C to validate the generalization.
We validate our method by regarding four different common corruptions, \ie, Gaussian noise (gn), pixelate (pl), fog (fg), and glass blur (gb), as the specific corruptions mentioned in \secref{subsec:problem}. Our objective is to enhance the robustness of an initial DNN to the four corruptions, respectively, while not harming the accuracy of the original testing dataset of CIFAR-10 and other corruption datasets. 

{\bf Metrics.}
In our experiment, we choose top-1 accuracy as the metric, that is, the proportion of the correct model judgment among the total number of testing images. 

{\bf Baseline methods.}
We consider the following baselines: \ding{182} To validate the effectiveness of the $K$-center greedy based $\mathcal{C}_\text{fail}^\text{xx}$ selection, we first set the random selection strategy as a baseline. Specifically, we randomly select $\mathcal{C}_\text{fail}^\text{xx}$ from $\mathcal{D}_\text{fail}^\text{xx}$ and perform the same steps as done in \secref{subsec:impl}.
Then, we can have two variants, \ie, RF-DARTS and RF-DARTSE, corresponding to our two versions, CF-DARTS and CF-DARTSE, respectively.
In addition, we augment training examples via a operation set (\ie, \{rotation, solarization, shear transformation, translation, auto-contrast adjustment, equalization\}) and randomly select 1000 examples to replace the failure examples $\mathcal{C}_\text{fail}$ in \reqref{eq:darts_q1_new} and \reqref{eq:darts_q2_new}, and conduct the same architecture search process as detailed in \secref{subsec:setups}. We get two variants denoted as the augmentation-set guided DARTS (AU-DARTS) and AU-DARTSE.
\ding{183} To validate the advantages of the proposed method over the popular data-augmentation-based robustness enhancement, we select the following state-of-the-art baseline methods:
CutMix \cite{Cutmix}, AugMix \cite{hendrycks2020augmix},  and DeepRepair \cite{yu2020deeprepair}. Specifically, CutMix augments training examples by cutting and pasting parts of the images while mixing the labels proportionally. AugMix stochastically samples a series of the operations from \{rotation, solarization, shear transformation, translation,
auto-contrast adjustment, equalization\} to augment the input image multiple times and conduct mix on transformed images. DeepRepair is an extension of the AugMix by adding a style-transfer method to the augmentation set, which uses the failure examples as the references to guide the transformations of the training data.
\ding{184} We also extend our method to more recent NAS methods and see whether the proposed method can benefit them and outperform their original versions. Specifically, we take PC-DARTS \cite{PC-DARTS}, DARTSPT \cite{ruochenwang2021dartspt}, and RobDARTS \cite{RobustDARTS}. PC-DARTS promotes searching efficiency, and DARTSPT focuses on architecture selection. In particular, RobDARTS is to enhance DARTS that yields degenerate architectures with very poor test performance.

%

\begin{table*}[t]
\centering
\setlength{\tabcolsep}{1pt}
\tiny
\caption{Top-1 accuracy of the initial DNN and refined DNNs with our CF-DARTS and CF-DARTSE on the original testing dataset (\ie, `Clean' column), the specific corruptions (\ie, Gauss$_\text{fail}$, Pixel$_\text{fail}$, and Glass$_\text{fail}$), and other 14 corruptions.}
\begin{tabular}{l|c|ccc|cccc|cccc|cccc}
\toprule 
\rowcolor{tabgray} &  & \multicolumn{3}{c|}{Noise} & \multicolumn{4}{c|}{Blur} & \multicolumn{4}{c|}{Weather} & \multicolumn{4}{c}{Digital}\tabularnewline
\midrule 
\rowcolor{tabgray} & Clean  & \cellcolor{tabhigh} Gauss$_\text{fail}$ & Shot  & Impulse  & Defocus  & Glass  & Motion  & Zoom  & Snow  & Frost  & Fog  & Bright  & Contrast  & Elastic  & Pixel  & Jpeg\tabularnewline
\midrule 
Org.~DNN  & 96.26  & \cellcolor{tabhigh} 0.00  & 45.18  & 51.53  & 87.49  & 56.43  & 82.11  & 81.71  & 87.58  & 80.46  & 91.30  & 94.96  & 82.05  & 88.31  & 76.30  & 82.67\tabularnewline
CF-DARTS  & 96.83   & \cellcolor{tabhigh} 18.27  & 50.21  & 49.25   & 87.74  & 68.03   & 84.96  & 82.67  & 89.35   & 84.67   & 92.08   & 95.66   & 83.69  & 89.92   & 78.96  & 84.04\tabularnewline
CF-DARTSE  & \textbf{97.35}  & \cellcolor{tabhigh} \textbf{87.01}  & \textbf{90.40}   & \textbf{75.93}   & \textbf{89.08}   & \textbf{76.26}  & \textbf{86.88}   & \textbf{84.21}   & \textbf{91.01}   & \textbf{89.66}  & \textbf{92.98}   & \textbf{96.39}  & \textbf{84.71}  & \textbf{91.00}   & \textbf{80.62}  & \textbf{85.28}\tabularnewline
\midrule 
\midrule 
\rowcolor{tabgray} & Clean  & Gauss  & Shot  & Impulse  & Defocus  & \cellcolor{tabhigh} Glass$_\text{fail}$ & Motion  & Zoom  & Snow  & Frost  & Fog  & Bright  & Contrast  & Elastic  & Pixel  & Jpeg\tabularnewline
\midrule 
Org. DNN  & 96.26  & 32.58  & 45.18  & 51.53  & 87.49   & \cellcolor{tabhigh} 0.00  & 82.11  & 81.71  & 87.58  & 80.46  & 91.30  & 94.96  & 82.05  & 88.31  & 76.30  & 82.67\tabularnewline
CF-DARTS  & 96.84  & 48.33  & 59.26   & 57.51   & 87.45   & \cellcolor{tabhigh} 49.86  & 85.22   & 80.85   & 89.88   & 84.71   & 91.82   & 95.80   & 83.56   & 89.94   & 79.44   & 84.33\tabularnewline
CF-DARTSE  & \textbf{97.15}  & \textbf{63.85}  & \textbf{69.19}  & \textbf{63.07}  & \textbf{90.74}  & \cellcolor{tabhigh} \textbf{82.06}  & \textbf{87.23}  & \textbf{85.78}  & \textbf{91.34}  & \textbf{90.73}  & \textbf{92.89}  & \textbf{95.88}  & \textbf{86.28}  & \textbf{91.43}  & \textbf{83.65}  & \textbf{84.95}\tabularnewline
\midrule 
\midrule 
\rowcolor{tabgray} & Clean  & Gauss  & Shot  & Impulse  & Defocus  & Glass & Motion  & Zoom  & Snow  & Frost  & Fog$_\text{fail}$  & Bright  & Contrast  & Elastic  & Pixel  & Jpeg\tabularnewline
\midrule 
Org. DNN  & 96.26  & 32.58  & 45.18  & 51.53  & 87.49   & 56.43  & 82.11  & 81.71  & 87.58  & 80.46  & \cellcolor{tabhigh} 0.00  & 94.96  & 82.05  & 88.31  & 76.30  & 82.67\tabularnewline
CF-DARTS  & 97.22  & \textbf{43.34}  & \textbf{55.14}   & 53.48   & 89.42   & \textbf{72.68}  & 86.39   & 84.01  & 90.70  & 87.40 & \cellcolor{tabhigh} 93.27   & 96.11 & 85.92  & 90.13  & 79.01  & 83.06 \tabularnewline
CF-DARTSE  & \textbf{99.32}  & 32.14  & 45.85  & \textbf{53.97}  & \textbf{92.67}  & 70.11  & \textbf{90.47}  & \textbf{88.20}  & \textbf{92.50}  & \textbf{88.45}  &  \cellcolor{tabhigh}  \textbf{97.05}  & \textbf{98.40}  & \textbf{92.77}  & \textbf{92.72}  & \textbf{79.88}  & \textbf{85.93}\tabularnewline
\midrule 
\midrule 
\rowcolor{tabgray} & Clean  & Gauss  & Shot  & Impulse  & Defocus  & Glass  & Motion  & Zoom  & Snow  & Frost  & Fog  & Bright  & Contrast  & Elastic  & \cellcolor{tabhigh} Pixel$_\text{fail}$ & Jpeg\tabularnewline
\midrule 
Org. DNN  & 96.26  & 32.58  & 45.18  & 51.53  & \textbf{87.49}  & 56.43  & 82.11  & 81.71  & 87.58  & 80.46  & 91.30  & 94.96  & \textbf{82.05}  & 88.31  & \cellcolor{tabhigh} 0.00  & 82.67\tabularnewline
CF-DARTS  & 96.55  & 39.45  & 52.73  & 52.03  & 86.79  & 60.50  & 83.00  & 82.10   & 88.62  & 83.91  & 91.35  & 95.22  & 81.19   & 88.50  & \cellcolor{tabhigh} 31.43  & 83.95\tabularnewline
CF-DARTSE  & \textbf{97.52}  & \textbf{43.80}  & \textbf{56.28}  & \textbf{55.01}  & 87.29  & \textbf{65.49}  & \textbf{84.12}  & \textbf{82.94}  & \textbf{90.53}  & \textbf{85.44}  & \textbf{91.65}  & \textbf{96.55}  & 81.52  & \textbf{89.94}  & \cellcolor{tabhigh} \textbf{87.62}  & \textbf{85.85}\tabularnewline
\bottomrule
\end{tabular}
\label{tab:pixel&glass}
\end{table*}

\subsection{Validation and Comparison Results} \label{subsec:valid}
{\bf Validation of proposed methods.} In this part, we validate our methods, \ie, CF-DARTS and CF-DARTSE, by taking the Gaussian noise (gn) as the specific corruption, \ie, `xx=gn' in \secref{subsec:impl}. 
To demonstrate the effectiveness of the core failure set selection method in Algorithm~\ref{alg}, we can simply replace the selected core-failure set with the randomly selected examples and augmented examples, thus get four baselines, \ie, RF-DARTS, RF-DARTSE, AU-DARTS, and AU-DARTSE (See more details in \secref{subsec:impl}).
Here, we report the accuracy on the original testing dataset (\ie, `Clean' column), $\mathcal{D}_\text{fail}^\text{gn}/\mathcal{C}_\text{fail}^\text{gn}$ (\ie, `Gauss$_\text{fail}$' column), and $14$ corrupted testing datasets in \tableref{tab:baseline}.  
All methods are evaluated five times and the averaged results, as well as standard deviation, are reported.

\begin{figure}[t]
	\centering
	\includegraphics[width=1.0\columnwidth]{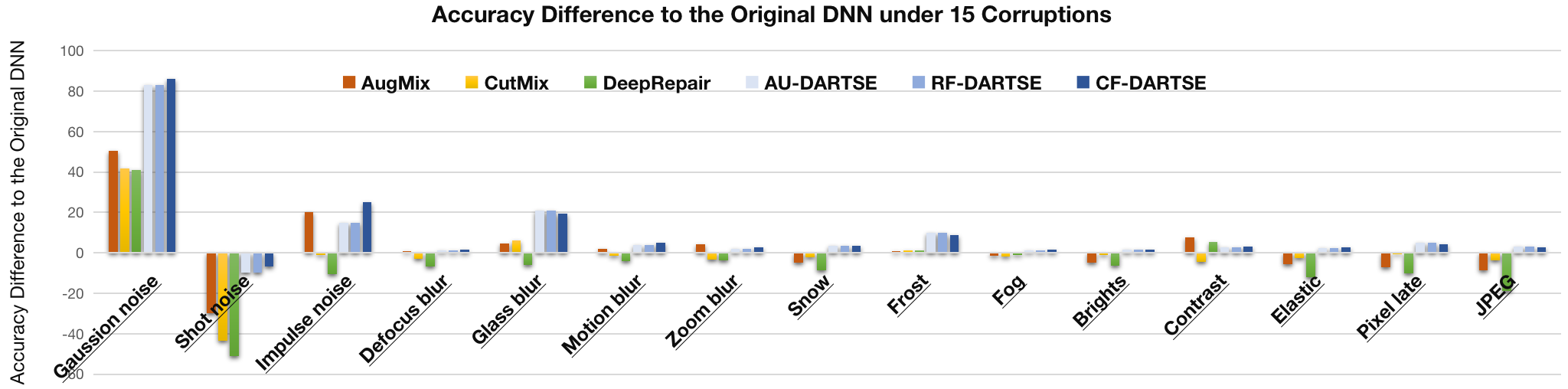}
	\caption{Accuracy difference between enhanced and original DNN via AugMix, CutMix, DeepRepair, AU-DARTSE, RF-DARTSE, and CF-DARTSE, respectively. That is, we use the scores of each row in \tableref{tab:baseline} to minus the first row.
	}
	\label{fig:diff}
\end{figure}

Moreover, we further present the accuracy variations of the DNNs searched by DARTS, CF-DARSTS, and CF-DARTSE during the training process. Specifically, we perform CF-DARTS, CF-DARTSE, and DARTS one by one on the same server and record the accuracy of the searched DNNs on the clean and corrupted testing datasets during the training process. Here, we consider three corruptions, \ie, Gaussian noise, shot noise, and impulse noise.
CF-DARTS and CF-DARTSE are guided by a few failure examples selected from $\mathcal{D}_\text{fail}^\text{gn}$. 
As shown in \figref{fig:budgecmp},  we see that the DNNs searched by our two methods (\ie, CF-DARTS and CF-DARTSE) outperform the one based on the basic DARTS along most of the training iterations under the three corruptions. Besides, the results of the DNNs on the clean dataset are almost the same. Overall, the experiments demonstrate that our two methods can enhance DNN's robustness significantly while not harming the capability on the clean data.

According to \tableref{tab:baseline}, we observe that: \ding{182} Our method CF-DARTS does enhance the robustness of the initial DNN to Gassusian noise since the accuracy on Gauss$_\text{fail}$ increases from $0.00\%$ to $18.27\%$. Moreover, the accuracy of the clean data and most of the other corruptions do not reduce and become even higher, demonstrating that the proposed method is able to enhance the robustness to a specific corruption while not harming the accuracy of clean data and robustness against other corruptions. \ding{183} CF-DARTSE significantly outperforms CF-DARTS on Gauss$_\text{fail}$ and also achieves much better scores on other datasets even if we just add 1,000 failure examples for retraining the refined architecture. It demonstrates that failure data itself benefits further robustness enhancement. \ding{184} CF-DARTS/DARTSE achieve better results across all cases than RF-DARTS/DARTSE and AU-DARTS/DARTSE on the Gauss$_\text{fail}$ and also achieve higher accuracy on most the other corrupted datasets. The results show that the selected failure examples can guide the robustness enhancement of the targeted DNN more effectively than randomly selected or augmented examples, which infers the effectiveness of our $K$-center greedy selection.

\begin{table*}[t]
\centering
\setlength{\tabcolsep}{1pt}
\renewcommand\arraystretch{1.2}
\tiny
\caption{Top-1 accuracy of the original DNN and refined DNNs on the raw Tiny-ImageNet testing dataset (\ie, `Clean' column) and respective 15 corruptions. Note that, the refined DNNs' architectures are searched on the training and validation datasets of CIFAR-10 and the selected failure set. RF-DARTS and CF-DARTS are our methods based on the random failure set selection and core failure set selection strategies, respectively.}
\begin{tabular}{ll|c|ccc|cccc|cccc|cccc}
\toprule 
\rowcolor{tabgray} \multicolumn{2}{c|}{} &  & \multicolumn{3}{c|}{Noise} & \multicolumn{4}{c|}{Blur} & \multicolumn{4}{c|}{Weather} & \multicolumn{4}{c}{Digital}\tabularnewline
\midrule 
\rowcolor{tabgray} \multicolumn{2}{c|}{} & Clean  & Gauss & Shot  & Impulse  & Defocus  & Glass  & Motion  & Zoom  & Snow  & Frost  & Fog  & Bright  & Contrast  & Elastic  & Pixel  & Jpeg \tabularnewline
\midrule 
\multicolumn{2}{c|}{Org.~DNN}  & 62.63 & \cellcolor{tabhigh} 23.55 & 29.06 & 24.14 & 22.44 & \cellcolor{tabhigh} 23.44 & 29.02 & 24.82 & 31.50 & 31.15 & \cellcolor{tabhigh} 28.94 & 39.29 & 13.53 & 35.63 & \cellcolor{tabhigh} 44.02 & 42.58\tabularnewline
\midrule 
\midrule 
\multirow{2}{*}{\makecell{Researched DNN \\ on $\mathcal{D}_\text{fail}^\text{gn}/\mathcal{C}_\text{fail}^\text{gn}$}} & RF-DARTS & 64.06 & \cellcolor{tabhigh} 25.09 & 30.44 & 25.42 & 21.75 &  24.17 & 28.00 & 25.29 & 31.30 & 31.93 & 29.69 & 38.47 & 13.99 & 36.32 & 44.90 & 44.07 \tabularnewline
 & CF-DARTS & 65.52 & \cellcolor{tabhigh}  25.43 & 31.46 & 26.35 & 23.47 & 23.00 & 29.83 & 26.36 & 32.99 & 33.86 & 32.04 & 40.88 & 14.92 & 37.62 & 46.38 & 45.10 \tabularnewline
\midrule
\midrule 
\multirow{2}{*}{\makecell{Researched DNN \\ on $\mathcal{D}_\text{fail}^\text{gb}/\mathcal{C}_\text{fail}^\text{gb}$}} & RF-DARTS & 64.26 & 25.75 & 30.95 & 25.84 & 21.66 & \cellcolor{tabhigh} 23.75 & 27.06 & 24.75 & 31.90 & 32.49 & 29.38 & 39.85 & 13.80 & 36.03 & 45.78 & 44.69 \tabularnewline
& CF-DARTS & 62.08 & 29.67 & 35.28 & 28.42 & 23.95 & \cellcolor{tabhigh} 28.28 & 33.12 & 27.79 & 36.09 & 36.02 & 31.57 & 42.83 & 15.73 & 41.50 & 48.21 & 50.20 \tabularnewline
\midrule 
\midrule 
\multirow{2}{*}{\makecell{Researched DNN \\ on $\mathcal{D}_\text{fail}^\text{fg}/\mathcal{C}_\text{fail}^\text{fg}$}} & RF-DARTS & 64.59 & 25.42 & 31.42 & 27.08 & 24.25 & 23.87 & 30.89 & 27.29 & 33.71 & 34.23 & \cellcolor{tabhigh} 30.97 & 41.19 & 14.50 & 37.52 & 46.10 & 45.38 \tabularnewline
& CF-DARTS & 66.00 & 26.14 & 32.15 & 26.90 & 24.31 & 23.72 & 29.89 & 26.83 & 33.19 & 34.56 & \cellcolor{tabhigh} 31.76 & 42.95 & 14.57 & 37.70 & 46.77 & 45.20 \tabularnewline
\midrule 
\midrule 
\multirow{2}{*}{\makecell{Researched DNN \\ on $\mathcal{D}_\text{fail}^\text{pl}/\mathcal{C}_\text{fail}^\text{pl}$}} & RF-DARTS & 63.40 & 23.95 & 29.86 & 24.72 & 23.14 & 22.54 & 29.09 & 25.81 & 30.77 & 30.77 & 29.19 & 39.75 & 14.11 & 35.42 & \cellcolor{tabhigh} 43.79 & 43.27\tabularnewline
& CF-DARTS & 64.04 & 25.13 & 30.37 & 26.02 & 23.10 & 23.17 & 29.07 & 25.28 & 31.72 & 32.72 & 29.91 & 40.91 & 14.19 & 35.70 & \cellcolor{tabhigh} 44.25 & 43.06 \tabularnewline
\bottomrule
\end{tabular}
\label{tab:generalization}
\end{table*}

{\bf Comparison to data-augmentation-based methods.} Following the experimental setups in \secref{subsec:valid}, we further compare our method with the pure data-augmentation-based robustness enhancements, \ie, DeepRepair \cite{yu2020deeprepair}, AugMix \cite{hendrycks2020augmix}, and CutMix \cite{Cutmix}. We use their default setups and retrain the initial DNN with their respective augmentation strategies. 
As presented in \tableref{tab:baseline} and \figref{fig:diff}, we see that: \ding{182} Compared with CF-DARTSE, DeepRepair, AugMix, and CutMix can also improve the robustness against the specific corruption, \ie, Gaussian noise, with the accuracy on Gauss$_\text{fail}$ from $0.00\%$ to $50.35\%$, $29.26\%$ and $21.99\%$, respectively. Nevertheless, they lead to slightly worse results on clean datasets and most the corrupted datasets. In particular, AugMix reduces the accuracy of clean images from $96.26\%$ to $89.75\%$. \ding{183} Our method CF-DARTSE outperforms the three methods significantly on most of the datasets with large margins, demonstrating that combining architecture refinement and failure example-enriched dataset can help enhance robustness better than data augmentation-based methods.

{\bf Comparison to other NAS methods.} We equip our method to more recent NAS methods and see whether the proposed method can benefit them and outperform their original versions. We follow the same process in \secref{subsec:impl}. Specifically, given a NAS method, we first search and train a DNN (\ie, $\phi_0$) and test it on different corrupted datasets. Then, given a few failure examples containing a specific corruption (\eg, Gaussian noise), we use our method to refine the architecture and enhance the robustness against the specific corruption. For each NAS method, we follow the setups in \secref{subsec:setups} and try four variants, \ie, RF-`**', RF-`**'E, CF-`**', and CF-`**'E where `**' denotes the name of the NAS method. \revised{
Specifically, we validate the effectiveness of our method on PC-DARTS \cite{PC-DARTS}, DARTSPT \cite{ruochenwang2021dartspt}, and RobDARTS \cite{RobustDARTS}, respectively.
As shown in \figref{tab:othernas}, for all of the three NAS methods, our methods (\ie, CF-`**' and CF-`**'E) enhance their robustness significantly and outperform the basic variants (\ie, RF-`**' and RF-`**'E) with large margins, which further demonstrates the effectiveness of the proposed framework.}

\begin{table*}[t]
\centering
\setlength{\tabcolsep}{1pt}
\tiny
\caption{Top-1 accuracy of the initial DNN and refined DNNs with RF-`**', RF-`**'E, CF-`**' and CF-`**'E on the original testing dataset (\ie, `Clean' column), the specific corruptions (\ie, Gauss$_\text{fail}$), and other 14 corruptions. `**' represents the names of different NAS methods, \ie, PC-DARTS \cite{PC-DARTS}, DARTSPT \cite{ruochenwang2021dartspt}, and RobDARTS \cite{RobustDARTS}.}
\begin{tabular}{l|c|ccc|cccc|cccc|cccc}
\toprule 
\rowcolor{tabgray} &  & \multicolumn{3}{c|}{Noise} & \multicolumn{4}{c|}{Blur} & \multicolumn{4}{c|}{Weather} & \multicolumn{4}{c}{Digital}\tabularnewline
\midrule 
\rowcolor{tabgray} & Clean  & \cellcolor{tabhigh} Gauss$_\text{fail}$ & Shot  & Impulse  & Defocus  & Glass  & Motion  & Zoom  & Snow  & Frost  & Fog  & Bright  & Contrast  & Elastic  & Pixel  & Jpeg\tabularnewline
\midrule 
PCDARTS & 97.00 & 0.00	& 49.38	& 58.08 & 84.24	& 67.69 & 79.86 & 79.00 & 90.01 & 85.78 & 92.01	& 95.63	& 85.42	& 86.90	& 73.39	& 81.22 \tabularnewline
RF-PCDARTS & 96.68 & 13.30 & 51.85	& 49.83	& 84.33	& 63.04	& 81.68	& 79.37	& 88.41	& 84.00 & 91.41 & 95.08 & 83.91 & 87.34 & 70.83 & 81.45\tabularnewline
RF-PCDARTSE & 97.01 & 84.32 & 88.05 & 73.28 & 85.00 & 71.55 & 82.04 & 79.90 & 89.86 & 88.08 & 92.06 & 95.94 & 83.15 & 88.08 & 74.15 & 82.39\tabularnewline
CF-PCDARTS & 96.84	& 13.76	& 52.60	& 51.53	& 84.96	& 64.90	& 81.66	& 80.03	& 88.75	& 84.97 & 91.61 & 95.51 & 83.33 & 87.46 & 71.80 & 80.99 \tabularnewline
CF-PCDARTSE & \textbf{96.95} & \textbf{88.69} & \textbf{90.48} & \textbf{77.66} & \textbf{85.20} & \textbf{72.65} & \textbf{82.45} & \textbf{81.20} & \textbf{89.86} & \textbf{88.86} & \textbf{92.59} & \textbf{96.83} & \textbf{84.84} & \textbf{88.30} & \textbf{74.32} & \textbf{82.98} \tabularnewline
\midrule
DARTSPT
& 97.41 & 0.00 & 48.23 & 54.18 & 89.81 & 76.03 & 86.06 & 84.07 & 90.32 & 86.16 & 92.26 & 96.25 & 86.59 & 90.29 & 80.53 & 83.59 \tabularnewline
RF-DARTSPT & 97.29 & 12.55 & 48.07 & 54.76 & 87.26 & 68.04 & 85.16 & 80.85 & 89.59 & 84.04 & 92.41 & 96.25 & 85.37 & 89.48 & 79.39 & 83.56 \tabularnewline
RF-DARTSPTE & 97.64 & 89.43	& 92.84	& 84.64	& 88.59	& 75.42	& 86.70	& 82.91	& 90.24 & 88.11 & 93.12 & 96.50 & 86.56 & 90.49 & 78.34 & 84.37 \tabularnewline
CF-DARTSPT & 97.32 & 14.74 & 48.80 & 56.25 & 88.12 & 65.35 & 85.73 & 82.35 & 90.14 & 84.60 & 92.85 & 96.12 & 86.57 & 89.88 & 79.57 & 83.45 \tabularnewline
CF-DARTSPTE & \textbf{97.75} & \textbf{90.75} & \textbf{92.85} & \textbf{86.12} & \textbf{90.09} & \textbf{75.89} & \textbf{87.29} & \textbf{84.98} & \textbf{91.24} & \textbf{88.49} & \textbf{93.55} & \textbf{96.59} & \textbf{86.83} & \textbf{90.88} & \textbf{83.64} & \textbf{85.97} \tabularnewline
\midrule
RobDARTS & 96.96 & 0.00 & 51.14 & 55.98 & 86.81 & 65.88 & 83.41 & 81.48 & 90.10 & 85.88 & 92.99 & 96.00 & 87.43 & 89.38 & 75.06 & 83.45 \tabularnewline
RF-RobDARTS & 97.20 & 6.83 & 41.95 & 46.16 & 88.24 & 68.23 & 86.44 & 82.30 & 89.40 & 84.75 & 92.36 & 95.85 & 85.41 & 90.22 & 80.40 & 84.11 \tabularnewline
RF-RobDARTSE & 97.32 & 87.86	& 91.14	& 84.77	& 89.13	& 74.49	& 86.69	& 82.84	& 90.81 &	87.42 & 92.44 & 96.33 & 85.51 & 90.64 & 81.51 & 85.62\tabularnewline
CF-RobDARTS & 97.30 & 12.73 & 48.95 & 52.15 & 88.73 & 72.21 & 86.29 & 82.85 & 90.39 & 85.77 & 93.25 & 95.97 & 85.40 & 90.39 & 77.14 & 83.85 \tabularnewline
%
CF-RobDARTSE & \textbf{97.88} & \textbf{88.55} & \textbf{91.80} & \textbf{86.49} & \textbf{90.23} & \textbf{77.42} & \textbf{87.43} & \textbf{83.99} & \textbf{91.66} & \textbf{89.08} & \textbf{93.68} & \textbf{96.75} & \textbf{86.92} & \textbf{91.51} & \textbf{81.98} & \textbf{85.73} \tabularnewline
\bottomrule
\end{tabular}
\label{tab:othernas}
\end{table*}


\subsection{Results of Different Corruptions}

In addition to the Gaussian noise corruption, we can further validate our method by considering other corruptions. However, it takes huge computing resources and training time to test all corruptions. Hence, we randomly select the other three corruptions from `Blur', `Digital', and `Weather' types for validation, \ie, the glass blur, pixelate, and fog corruptions. The results are reported in \tableref{tab:pixel&glass}.  
The results further demonstrate the effectiveness of our method: \ding{182} Both methods, \ie, CF-DARTS and CF-DARTSE,  on all three corruptions let refined DNNs outperform the original DNN significantly on all datasets including the clean testing dataset and other $14$ corrupted datasets. \ding{183} CF-DARTSE using the 1000 failure examples to retrain the refined architectures increases the accuracy on all datasets significantly.

\subsection{Generalization to Tiny-ImageNet/Tiny-ImageNet-C} 
In the subsection, we aim to validate whether the neural networks searched on the training dataset of CIFAR-10 and the selected core-failure examples can generalize to other datasets, \ie, Tiny-ImageNet and Tiny-ImageNet-C. 
Specifically, as the method introduced in \secref{subsec:cfdarts}, we can get refined architecture by searching on the training and validation datasets of CIFAR-10 (\ie, $\mathcal{D}_\text{train}$ and $\mathcal{D}_\text{val}$) and the selected core-failure-set (\ie, $\mathcal{C}^{\text{xx}}_\text{fail-t}$ and $\mathcal{C}^{\text{xx}}_\text{fail-v}$) where $\text{xx}$ denotes the name short for specific corruptions.
Here, we retrain the searched architecture based on the training dataset of Tiny-ImageNet. Then, we further evaluate the searched architecture on the testing dataset of Tiny-ImageNet and its fifteen corrupted datasets.
We report the results in \tableref{tab:generalization} and have the following observations: \ding{182} Compared with the original DNN, the CF-DARTS-searched architectures achieve much higher accuracy on both clean and corrupted datasets the most times. For example, the architecture searched on $\mathcal{D}_\text{train}$ and $\mathcal{D}_\text{val}$ of CIFAR-10 and the selected core-failure-set $\mathcal{C}^{\text{gn}}_\text{fail}$ under Gaussian noise (gn) has the accuracy of 65.52\% on the raw Tiny-ImageNet dataset (\ie, the `Clean' column), which is significantly higher than the original DNN with the accuracy of 62.63\%. Moreover, it also achieves much higher accuracy on all corrupted datasets except the glass blur. We have similar observations on other searched architectures under pixelate, fog, and glass blur. \ding{183} Compared with the architecture searched by baseline method RF-DARTS, CF-DARTS gets higher accuracy on most of the corrupted datasets. For example, for the architecture searched with the core-failure-set under Gaussian noise (gn) corruption, CF-DARTS outperforms RF-DARTS on the clean and fifteen corrupted datasets. 
Overall, the experiments demonstrate that our method can search architectures with high generalizations, which can generalize to a totally different dataset.

\newcolumntype{g}{>{\columncolor{tabhigh}}c}
\begin{table*}[t]
\centering
\setlength{\tabcolsep}{1pt}
\tiny
\caption{ Top: Top-1 accuracy of the initial DNN and refined DNNs via RF-DARTS and CF-DARTS with different sizes of $\mathcal{C}_\text{fail}$ on the original testing dataset (\ie, `Clean' column), the specific corruptions (\ie, Gauss$_\text{fail}$), and other 14 corruptions. Bottom: Top-1 accuracy of another sophisticated pre-trained DNN (\ie `Org.~DNN-v2') and its refined counterparts through our RF-DARTS and CF-DARTS on the original testing dataset (\ie, `Clean' column), the specific corruptions (\ie, Gauss$_\text{fail}$), and other 14 corruptions.}
\begin{tabular}{l|c|gcc|cccc|cccc|cccc}
\toprule 
\rowcolor{tabgray} &  & \multicolumn{3}{c|}{Noise} & \multicolumn{4}{c|}{Blur} & \multicolumn{4}{c|}{Weather} & \multicolumn{4}{c}{Digital}\tabularnewline
\midrule 
\rowcolor{tabgray} & Clean  & Gauss$_\text{fail}$ & Shot  & Impulse  & Defocus  & Glass  & Motion  & Zoom  & Snow  & Frost  & Fog  & Bright  & Contrast  & Elastic  & Pixel  & Jpeg\tabularnewline
\midrule 
Org. DNN  & 96.26  & 0.00  & 45.18  & 51.53  & 87.49  & 56.43  & 82.11  & 81.71  & 87.58  & 80.46  & 91.30  & 94.96  & 82.05  & 88.31  & 76.30  & 82.67\tabularnewline
\midrule 
RF-DARTS(1000)  & 96.73  & 14.06  & 43.94  & 56.96  & 86.48  & 66.47  & 85.24  & 79.33   & 87.88  & 82.02  & 92.03  & 95.47  & 85.17  & 89.75  & 77.79  & 83.40\tabularnewline
 CF-DARTS(1000)  & 96.62  & \textbf{20.05}  & 53.55  & 46.09  & 87.67  & 71.93  & 84.93  & 81.43   & 89.11  & 85.26  & 91.62  & 95.41  & 84.51  & 89.96  & 78.23  & 84.55\tabularnewline
\midrule 
RF-DARTS(5000) & 96.58  & 15.16  & 46.73  & 51.53  & 86.36  & 65.85  & 84.44  & 80.68   & 87.36   & 82.27  & 91.50  & 95.23  & 84.41  & 89.46  & 77.13  & 83.74\tabularnewline
 CF-DARTS(5000)  & 96.53  & \textbf{22.37}  & 53.76  & 48.24  & 84.32   & 66.72  & 82.81  & 75.54   & 87.64  & 82.53  & 91.36  & 95.26  & 82.33  & 89.19  & 78.07  & 84.30\tabularnewline
\midrule 
RF-DARTS(10000) & 96.77  & 11.82  & 46.97  & 51.20   & 87.93  & 65.81  & 85.03  & 83.41  & 89.02  & 84.13  & 91.96  & 95.39  & 82.67  & 89.85  & 80.61  & 83.43\tabularnewline
 CF-DARTS(10000)  & 96.76  & \textbf{16.59}  & 53.72  & 50.16  & 87.59  & 67.71  & 84.07  & 82.21  & 89.63  & 85.22  & 91.70  & 95.76  & 82.61  & 89.59  & 77.03  & 84.02\tabularnewline
\midrule 
RF-DARTS(15000) & 96.50  & 10.25  & 44.75  & 54.16  & 85.37  & 67.35  & 82.56  & 78.70   & 87.65  & 81.94  & 90.65  & 94.95   & 81.94   & 88.69  & 78.04  & 83.80\tabularnewline
 CF-DARTS(15000) & 96.93  & \textbf{14.46}  & 51.82  & 55.26  & 88.46  & 70.39  & 85.63  & 82.14  & 89.00  & 83.28  & 91.90  & 95.69  & 84.88  & 90.05  & 79.30  & 83.37\tabularnewline
\midrule 
RF-DARTS(20000) & 96.97  & 12.12  & 49.52  & 50.07  & 88.90  & 67.40  & 87.01  & 84.20  & 89.44  & 84.61  & 92.33  & 95.82  & 83.58  & 90.26  & 78.12  & 84.66\tabularnewline
 CF-DARTS(20000) & 96.51  & \textbf{14.21}  & 55.70  & 51.85  & 87.58  & 69.35  & 85.75  & 80.83   & 88.58  & 82.92  & 91.44  & 95.42  & 83.30  & 89.61  & 77.99  & 84.02\tabularnewline
\midrule 
\midrule 
Org.~DNN-v2  & 97.36  & \cellcolor{tabhigh} 0.00  & 47.21  & 54.43  & 87.81  & 66.25  & 84.30  & 82.18  & 89.41  & 84.17  & 92.63  & 96.17  & 84.23  & 89.51  & 78.16  & 83.53\tabularnewline
\midrule 
RF-DARTS  & 95.91   & \cellcolor{tabhigh} 18.35  & 49.82  & 49.34   & 83.09  & 63.82   & 78.70  & 77.90  & 86.61   & 82.78   & 91.06   & 94.64   & 82.58  & 86.17   & 70.74  & 80.36\tabularnewline
CF-DARTS  & 96.24  & \cellcolor{tabhigh} \textbf{19.12}  & 51.72   & 59.53   & 83.28   & 65.13  & 79.60   & 78.77   & 86.84   & 82.93  & 90.72   & 94.81  & 79.51  & 87.40   & 74.06  & 81.55\tabularnewline
\midrule 
RF-DARTSE  & 95.95  & \cellcolor{tabhigh} 73.07  & 81.81   & 65.57   & 82.80   & 66.64  & 79.29   & 78.36   & 87.91   & 86.09  & 90.10   & 94.75  & 80.19  & 86.29   & 73.27  & 80.89\tabularnewline
CF-DARTSE  & 96.40  & \cellcolor{tabhigh} \textbf{83.25}  & 88.07   & 75.41   & 85.17   & 72.95  & 82.39   & 80.95   & 88.31   & 88.02  & 90.76   &  95.10 & 82.45  & 88.18   & 74.61  & 83.16\tabularnewline
\bottomrule
\end{tabular}
\label{tab:different_number}
\end{table*}

\subsection{Influence of Failure Examples}
\label{subsec:influence_of_failure}

In previous experiments, we fix the size of core failure set $\mathcal{C}_\text{fail}$ as 1,000 examples (\ie, $B=1,000$). Here, we further discuss its influence by setting $B=\{1,000, 5,000, 10,000, 15,000, 20,000\}$ and report the results of RF-DARTS and CF-DARTS in \tableref{tab:different_number} (Top). We have the following observations: \ding{182} the accuracy on all datasets does not increase as the size of $\mathcal{C}_\text{fail}$ becomes larger, which is different from the common understanding of training model weights where the accuracy keeps a positive relationship with the size of the training dataset, since we just use $\mathcal{C}_\text{fail}$ for architecture refinement but not for model weight retraining.
\ding{183} although the size of $\mathcal{C}_\text{fail}$ is different, CF-DARTS always outperforms RF-DARTS, further demonstrating the effectiveness of the proposed core-failure-set selection method.


\subsection{Influences of Initial Networks}

In our experiment, we found that the architecture searched by the original DARTS does not always perform well. To eliminate the possible impact of deviation of the accuracy of the initial network, we generated five additional architectures by DARTS and selected the best model as our `Org.~DNN-v2'.
%
%
As shown in \tableref{tab:different_number} (Bottom), compared with the `Org.~DNN', the new version with more refinements achieved much better accuracy on the clean testing dataset and 14 corrupted datasets. Even then, our methods (\ie, CF-DARTS and CF-DARTSE) still enhance the robustness against the specific corruption (\ie, Gauss$_\text{fail}$) and other 14 corruptions significantly with obvious advantages over RF-DARTS and RF-DARTSE.


\subsection{Influences of Iteration Numbers}
\label{subsec:iteration}

\begin{figure}[t]
	\centering
	\includegraphics[width=1.0\columnwidth]{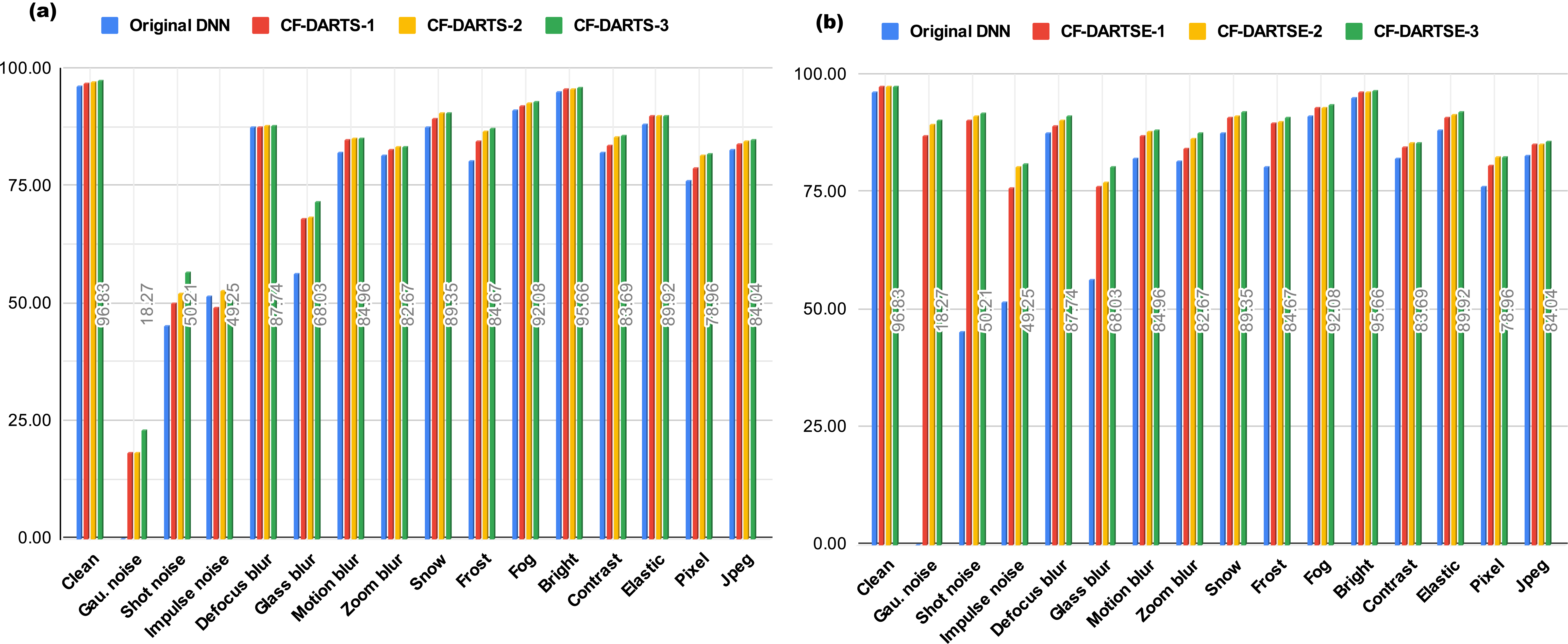}
	\caption{Accuracy comparison of CF-DARTS(E)-\{1,2,3\} and the original DNN. CF-DARTS(E)-\{1,2,3\} are used the refine the $\phi_0$ with with the guidance of Gaussian noise corruption. (a) displays the results of CF-DARTS-\{1,2,3\} and (b) represents the results of CF-DARTSE-\{1,2,3\}.
	}
	\label{fig:iterations}
\end{figure}

As detailed in \secref{subsec:impl}, our method can be conducted for several iterations where each iteration contains the optimization of the architecture and weights with 20 epochs. In the previous experiments, we set the iteration number as one. Here, we study the influence of the iteration numbers beyond one by taking the Gaussian noise (gn) as the specific corruption (\ie, `xx=gn' in \secref{subsec:impl}). Specifically, we try the CF-DARTS and CF-DARTSE for robustness enhancement of the original DNN $\phi_0$ with the iteration numbers as 1, 2, and 3, respectively, which are denoted as CF-DARTS(E)-\{1,2,3\}. We show the results in \figref{fig:iterations} and see that: \textit{First}, all CF-DARTS methods enhance the robustness of $\phi_0$ against the Gaussian noise as well as other corruption types. \textit{Second}, all CF-DARTS methods do not harm the accuracy of the clean images. \textit{Third}, the accuracy of enhanced DNNs becomes higher as the iteration number is larger under almost all corruptions.

\begin{table*}[t]
\centering
\tiny
\caption{Time cost of CF-DARTS.}
\begin{tabular}{l|r|ccc}
\toprule 
\rowcolor{tabgray} & Total & Core-Failure-Set Search & Network Architecture Search & Network Retraining \tabularnewline
\midrule 
Avg. Time (Hours) & 44.1472  & 0.4732   & 12.1700 & 31.5040 \\
Std. Dev.         & 1.5049   & 0.0314   & 0.0935  & 1.4598 \\
Ratio             & 100.0~\% & 1.0~\%   & 27.6~\% & 71.4~\%\\
\bottomrule
\end{tabular}
\label{tab:timeanalysis}
\end{table*}

\subsection{Cost Analysis}
\label{subsec:time}
As detailed in \secref{subsec:impl}, our method consists of three processes, that is, core-failure set search, network architecture search, and network retraining. We report the average time cost of the whole process in \tableref{tab:timeanalysis} and observe that the core-failure-set search only accounts for 1.0\% of the whole cost while the network architecture search and retraining take 99.0\% time cost, which demonstrates that our method has a limited effect on the time cost of the original DARTS.


\section{Conclusion}

In this paper, we have investigated how to refine a deployed DNN model's architecture for enhancing its robustness with the guidance of a few collected and misclassified examples that are degraded by some unknown but specific corruption patterns in the wild.
We have made a surprising and interesting observation that by merely adding a few corrupted and misclassified examples into the clean training dataset, we can already refine the model architecture and significantly enhance the model robustness. 
We have further proposed a novel \textit{core-failure-set guided DARTS} that embeds a $K$-center-greedy algorithm for DARTS to select suitable corrupted failure examples to refine the model architecture. Compared with the raw NAS method as well as the SOTA data-augmentation-based enhancement methods, our final method can achieve higher accuracy on both corrupted datasets and the original clean dataset. In particular, on some of the corruptions, we can achieve over $45\%$ absolute accuracy improvements. 
Although achieving progress, the method does not fully utilize the failure patterns in collected failure examples and deliveries them to the training data. In the future, we will further extend the proposed method by combining it with the data augmentation methods to fill the gap. Moreover, we could extend the method to video data and other visual intelligent tasks.

\section*{Acknowledgement}

This work was supported in part by the National Natural Science Foundation of China (61906135, 62020106004, 92048301); and in part by the Tianjin Science and Technology Plan Project (20JCQNJC01350). This research was also supported in part by Canada CIFAR AI Chairs Program, Amii RAP program, the Natural Sciences and Engineering Research Council of Canada (NSERC No.RGPIN-2021-02549, No.RGPAS-2021-00034, No.DGECR-2021-00019), as well as JSPS KAKENHI Grant No.JP20H04168, No.JP21H04877, JST-Mirai Program Grant No.JPMJMI20B8.

 \bibliographystyle{elsarticle-num} 
 \bibliography{cas-refs}

\newpage
\pagestyle{empty}

\end{document}